\documentclass[11pt]{article}

\usepackage{EMNLP2023}

\usepackage{times}
\usepackage{latexsym}

\usepackage[T1]{fontenc}
\usepackage[utf8]{inputenc}

\usepackage{microtype}

\usepackage{inconsolata}

\usepackage{graphicx}
\usepackage{amsmath}
\usepackage{amsthm}
\usepackage{booktabs}
\usepackage[ruled,linesnumbered]{algorithm2e}
\usepackage{array}
\usepackage{multirow}
\usepackage{setspace}
\usepackage{subcaption}
\usepackage{bm}

\newcolumntype{P}[1]{>{\centering\arraybackslash}p{#1}}

\title{Inspire the Large Language Model by External Knowledge on \\ BioMedical Named Entity Recognition}

\author{Junyi Bian\textsuperscript{\rm 1 \rm 3}, Jiaxuan Zheng\textsuperscript{\rm 2}, Yuyi Zhang\textsuperscript{\rm 1},
{\bf Shanfeng Zhu} \textsuperscript{\rm 2} \\
\textsuperscript{\rm 1} School of Computer Science, Fudan University, Shanghai 200433, China\\
\textsuperscript{\rm 2} Institute of Science and Technology for Brain-Inspired Intelligence, Fudan University, China\\
\textsuperscript{\rm 3} Shanghai Key Lab of Intelligent Information Processing, \\ Fudan University, Shanghai 200433, China\\
\texttt{\{zhusf, 20110240003\}@fudan.edu.cn} \\
}

\begin{document}
\maketitle

\begin{abstract}
Large language models (LLMs) have demonstrated dominating performance in many NLP tasks, especially on generative tasks. However, they often fall short in some information extraction tasks, particularly those requiring domain-specific knowledge, such as Biomedical Named Entity Recognition (NER). In this paper, inspired by Chain-of-thought, we leverage the LLM to solve the Biomedical NER step-by-step: break down the NER task into entity span extraction and entity type determination. Additionly, for entity type determination, we inject entity knowledge to address the problem that LLM's lack of domain knowledge when predicting entity category. Experimental results show a significant improvement in our two-step BioNER approach compared to previous few-shot LLM baseline. Additionally, the incorporation of external knowledge significantly enhances entity category determination performance.
\end{abstract}

% 大纲：
%   简单介绍一下 BNER
%   描绘一下大模型在 BNER 里面的应用
%   说明一下目前大模型在 BNER 中会出现的问题 （其中一个问题是LLM缺乏专业领域的知识）
%   这里就开始介绍我们的方法，我们的方法把 BNER 分成两个步骤解决，第一步是抽取实体span，第二步是对实体的 span 进行判定，这个 pipeline 类似 chain-of-thought。另外我们的方法在第二步引入了外部知识来增强模型的性能。获取知识的方法是从知识库之中进行实体的检索，同时我们还设计了一个无监督的方法从 UMLS 里构建知识库。
%   我们的贡献：1. 设计了一种引入外部知识来增强LLM的方法 2. 我们设计了一个无监督的方法来构建匹配知识库 3. 在多个生物医学NER数据集上进行实验达到了 unsupervised 的最好结果

\section{Introduction}

% 生物医学命名实体识别（BioNER）主要任务是提取出文本中具有特殊意义的biomedical实体，如疾病、基因、蛋白质等，相比于通用领域的 NER 任务，BioNER 任务数据集更多，实体名称更复杂，传统的solution是使用 BiLSTM-CRF based method 和 BERT based method 。

Biomedical Named Entity Recognition (BioNER) primarily focuses on extracting biomedical entities with specific significance from text, such as diseases, genes, proteins, etc. Compared to general domain NER tasks, BioNER datasets are more extensive, and entity names are more complex. Tradictional solutions often involve the use of BiLSTM-CRF based methods and BERT-based methods.

% 最近，大语言模型在多个 NLP 领域展现出了惊人的效果，在一些生成式的 NLP 任务上，例如 machine translation，question answer，结合 in-context learning，few-shot learning 等技术，就能取得比之前 supervised 方法更好的效果。

Recently, large language models have shown remarkable performance in various NLP domains. In some generative NLP tasks such as machine translation, question answering, combined with in-context learning, chain-of-thought, and other techniques, they can even achieve better results than previous supervised methods.

% 然而 llm 在抽取实体任务上的效果相比于其他生成式的任务并不理想。其中主要原因是实体抽取的范式和llm的预训练范式不太统一。受到“链式思维”（chain-of-thought）方法的启发，即把一个复杂的任务拆解成多个简单的生成式任务让llm来回答，我们也把ner分成两个步骤，第一步是使用 LLM 只负责对实体边界进行抽取，第二步再调用 LLM 来判断实体的类别。分两步进行，除了简化了NER任务，在第二步还有一个纠错的机会，也可以缓解第一步可能碰到的 hallucination issue。

However, the performance of large language models (LLM) in entity extraction tasks is not as promising as in other generative tasks. The primary reason for this discrepancy lies in the mismatch between the entity extraction paradigm and the pretraining paradigm of LLMs. Inspired by the 'chain-of-thought' method, which involves breaking down a complex task into multiple simpler generative tasks for LLMs to answer, we also split NER into two steps. The first step involves using LLM solely to extract entity boundaries, and in the second step, we invoke LLM again to determine the entity's category. This two-step approach not only simplifies the NER task but also provides an error-correction opportunity in the second step, helping alleviate potential hallucination issues that may arise in the first step.

% 但是即便如此，在特定领域的实体抽取上，LLM 的效果还是表现的还不是很好，例如在 biomedical 领域。从人类的角度来看，判断一个单词或者短语是否是一个实体可以通过一些经验或者分析语法来估计，但是判断一个实体的类别尤其是用缩写表示的实体，即使结合上下文也很难进行判断。所以，如果llm在训练预料中没有接触过领域的知识，很难从语义和上下文信息中推断出实体类别。我们的做法是引入一些外部的知识来辅助 LLM 进行实体的类别判断。More specific，对于一个实体名称，我们会从 UMLS 知识库中检索出和这个名称相似的一些实体和对应的信息，来辅助 LLM 进一步更精确的判断实体。

However, even with this improvement, the performance of LLMs in entity extraction tasks, especially in specific domains like biomedical, remains less than satisfactory. From a human perspective, deciding whether a word or phrase is an entity can often be estimated through experience or grammatical analysis. However, determining the category of an entity, especially when it's represented by abbreviations, is challenging even when considering context. Therefore, if LLMs have not been exposed to domain-specific knowledge during training, inferring entity categories from semantics and contextual information is difficult. Our approach involves introducing external knowledge to assist LLMs in entity category determination. More specifically, for an entity name, we retrieve similar entities and corresponding information from the UMLS knowledge base to aid LLMs in making more accurate determinations.

% in this paper，我们的贡献如下：1）利用 LLM ，我们采用先抽取实体类别，再进行实体类别的判断的流程来提升 LLM based NER performance 2）设计了一种引入外部知识来增强 LLM 实体类别判断的方法

In this paper, our contributions are as follows: 1) Leveraging LLM, we enhance LLM-based NER performance by following a process of first extracting entity boundaries and then determining entity categories. 2) We have designed a method for incorporating external knowledge to augment LLM-based entity category determination.

\section{Related Work}
% [在NER任务的早期阶段，研究人员主要依赖手工编写的规则和词典来识别命名实体。这些方法通过对文本进行模式匹配来寻找特定的实体名称。然而，这些方法在处理复杂的文本和大规模数据时存在限制。随着机器学习技术的发展，NER任务开始采用了基于统计和机器学习的方法，其中包括马尔可夫模型（Hidden Markov Model，HMM）和条件随机场（Conditional Random Fields，CRF）。]

%因为文献摘要都是英文的，所以直接用英文，改变一下句子结构比较方便，所以没有用中文写。主要就是罗列相关的NER文章的主要方法和成果。

%Named Entity Recognition (NER) is a task to identify key information in the text and classify it into a set of predefined categories,which is used for instance for identifying the names of persons, locations and organizations in text.
% \cite{2019_unified-MRC} proposed a unified framework and formulate the task of NER as a machine reading comprehension (MRC) task.\cite{2019_TENER} proposed a NER architecture adopting adapted Transformer Encoder to model the character-level and word-level features by incorporating the direction aware, distance-aware and un-scaled attention.\cite{2020_Luke} proposed new pretrained contextualized representations of words and entities based on the bidirectional transformer ,as well as an extension of the self-attention mechanism of the transformer.\cite{2020_Pyramid} proposed a layered model for nested NER.\cite{2020_Automated} proposed ACE to automate the process of finding better concatenations of embeddings for structured prediction tasks.

\subsection{Named Entity Recognition}
In the early stages of NER tasks, researchers primarily relied on manually crafted rules and dictionaries to identify named entities. With the rise of deep learning, NER tasks are commonly regarded as end-to-end sequence labeling tasks, with classic LSTM-CRF model\cite{2016_End-to-end} and BERT-based model \cite{2019_bert, 2020_biobert}. However, in recent years, the paradigm of NER has gradually shifted towards generative models\cite{2022_uie}. With the advent of large language models, both supervised\cite{2023_uniner} and unsupervised\cite{2023_gptner} methods have achieved significant results.

% 但是近几年，NER 的范式逐渐转向生成式模型，随着最近大模型的到来，一些无监督和有监督的党法都取得了显著的成果。

% 上面这些方法大部分把NER问题当成 one-step problem 来解决，而我们的策略是将 NER 当成 two-step problem 来做。
Most of the methods mentioned above approach the NER problem as an end-to-end problem, whereas our strategy is to treat NER as a two-step problem.
Some span-based NER methods also handle NER in two steps, where DMNER\cite{2023_dmner} are relative to our work, they treat NER as entity extraction and entity linking. However, DMNER uses a dictionary for linking entities constructed from the validation set and training set, and the entity extraction is borrowed from existing supervised NER models. In contrast, we use LLM for few-shot learning for both entity extraction and linking.

\subsection{Biomedical Named Entity Recognition}
BioMedical NER tries to automatically recognize biomedical entities in natural language text.\cite{2019_Cross} proposed a multi-task learning framework for BNER to collectively use the training data of different types of entities and improved the performance on each of them.\cite{2019_pubtator} improved concept identification systems and a new disambiguation module based on deep learning increase annotation accuracy.\cite{2020_huner} proposed the stand-alone NER tool HUNER incorporating fully trained models for five entity types.\cite{2021_hunflair} proposed HunFlair, a NER tagger that is easy to use, covers multiple entity types, is highly accurate and is robust toward variations in text genre and style. \cite{2023_aioner} proposed a novel all-in-one (AIO) scheme that uses external data from existing annotated resources to enhance the accuracy and stability of BioNER models.

\subsection{Knowledge-Enhanced LLM}
%[大语言模型的使用在自然语言处理中已经成为常规，因为其在生成文本和推理任务中表现良好。但是LLMs其中一个致命的缺点是缺少事实性的准确率，这是因为大语言模型有时会为了回答而捏造事实。因此，为了提升LLMs的可信度和真实性， Chain-of-Thought (CoT)\cite{2022_COT} 被提出来。CoT可以处理相对复杂的推理问题，例如数学题、常识推理等，同时也可以生成可被人类解度的推理思路，由此提升了LLMs的真实性。之后有许多团队都在探索如何去利用CoT来做出更好的预测结果。但是绝大部分的团队注重于单纯使用CoT来增强实验表现，只有少部分团队在研究如何提升CoT方法本身的质量。（如需具体例子详见Verify and Edit)]

%[要提升CoT的质量首先要厘清它的缺陷在哪里。之前提到过事实正确性一直是大语言模型的痛点，这是因为在微调阶段事实的信息来源是接触不到的，同时作为知识基底时，大语言模型无法回忆起精确的细节。因此，为了更好的控制文本生成和预测的事实正确性，从外部引入知识可以帮助大语言模型做出更好、更真实的回答和预测。\cite{2023_verify} 提出Verify-and-Edit (VE) 来后编辑推理链从而获得更加事实相关的预测。\cite{2023_CoK} 提出Chain of Knowledge (CoK) ，采用结构化知识库增强大型语言模型以提高事实正确性和减少事实捏造。]
The use of large language models (LLMs) has become commonplace in natural language processing due to their strong performance in text generation and inference tasks. However, one of the fatal flaws of LLMs is their lack of factual accuracy, as they sometimes generate fabricated facts to answer questions. Therefore, to enhance the credibility and truthfulness of LLMs, Chain-of-Thought (CoT)\cite{2022_COT} has been proposed. CoT can handle relatively complex reasoning problems, such as mathematical questions and common-sense reasoning. It can also generate reasoning paths that are interpretable by humans, thus improving the factual correctness of LLMs. Subsequently, many teams have been exploring how to utilize CoT to make better prediction results. However, the majority of teams have focused on simply using CoT to enhance experimental performance, with only a few teams researching how to improve the quality of the CoT method itself.
\medskip
% related work 这一块有点讲的太详细了，其实平均下来，每篇被引用的论文 1-2 句话就带过了
To improve the quality of CoT, it is first necessary to identify its shortcomings. As mentioned earlier, factual correctness has always been a pain point for large language models because the source of factual information during fine-tuning is inaccessible, and as knowledge bases, large language models cannot recall precise details. Therefore, to better control the factual correctness of text generation and predictions, introducing external knowledge can help large language models provide better and more accurate answers and predictions. Verify-and-Edit (VE) was proposed in \cite{2023_verify} to post-edit reasoning chains and obtain more factually relevant predictions. Chain of Knowledge (CoK) was proposed in \cite{2023_CoK}, using structured knowledge bases to enhance large language models to improve factual correctness and reduce the generation of fabricated facts.

% 我们方法和 GPTNER 的区别：实体抽取的
% 我们方法和 DMNER 的区别

% 大纲：
%   首先介绍 overall framework
%   
%   

\section{Methodology}
\subsection{Overall Framework}
% Follow COT，我们的方法把bner当成两个步骤来解决，这个思想和cot很接近。在第一步中，我们需要确定每个实体的边界

% [在第一步中，我们需要确定每个实体的边界，这个过程称为实体边界检测（Entity Boundary Detection，EBD）。这使我们能够精确地定位文本中的实体，并为下一步的实体分类提供清晰的边界。在该模块中，我们借鉴了GPT-NER的实现方法，这将在第3.2部分详细阐述。]

% [在第二步中，我们再次使用ChatGPT来预测每个提取出的实体的类别，称为实体类型预测（Entity Type Prediction，ETP）。通过ETP，我们可以为每个实体分配准确的类别标签，从而完成命名实体识别任务。]
% [值得注意的是，为了让我们的NER模型能够在特定领域如生物医学研究中表现出更好的性能，本文采用了一种无监督方法来构建基于UMLS数据集的外部知识库。我们将这个外部知识库整合到ETP模块中，以增强整个系统的性能。这种方法使我们的模型能够更好地适应生物医学领域的语境，理解上下文信息，并更准确地进行实体分类。这也是本文的主要贡献之一。]

We decompose the BNER process into two distinct steps.
In the first step, we need to determine the span boundaries of each entity. This enables us to precisely locate entities within the text and provide clear boundaries for the subsequent entity classification. In this module, we draw upon the implementation approach of GPT-NER\cite{2023_gptner}, which will be elaborated on in Section 3.2.

In the second step, we once again employ ChatGPT to predict the category of each extracted entity, and we  refer to this process as Entity Type Prediction (ETP). Through ETP, we can assign accurate category labels to each entity, thereby accomplishing the named entity recognition task.

\subsection{Entity Span Extraction}
% Follow GPT-NER，我们使用“@@##”这样的特殊标记来标出实体位置。例如，在输入文本“A common feature of these proteins is involvement with heterochromatin and/or transcriptional repression”中查找“Chemicals”类别实体，就可以被转换成生成文本序列“A common feature of these @@proteins## is involvement with heterochromatin and/or transcriptional repression”的任务。同时也结合 in-context learning，我们在输入给出 2 个例子，这两个例子主要是告诉模型输出的格式是什么。

% 在LLM抽取实体的时候，输入的prompt会指定实体的类别。对于包含多个类别的数据集，每次让LLM抽取一种类型的实体，然后不考虑实体的类别，把得到的span合并。

\begin{figure}[t]
    \centering
    \includegraphics[width=1\columnwidth]{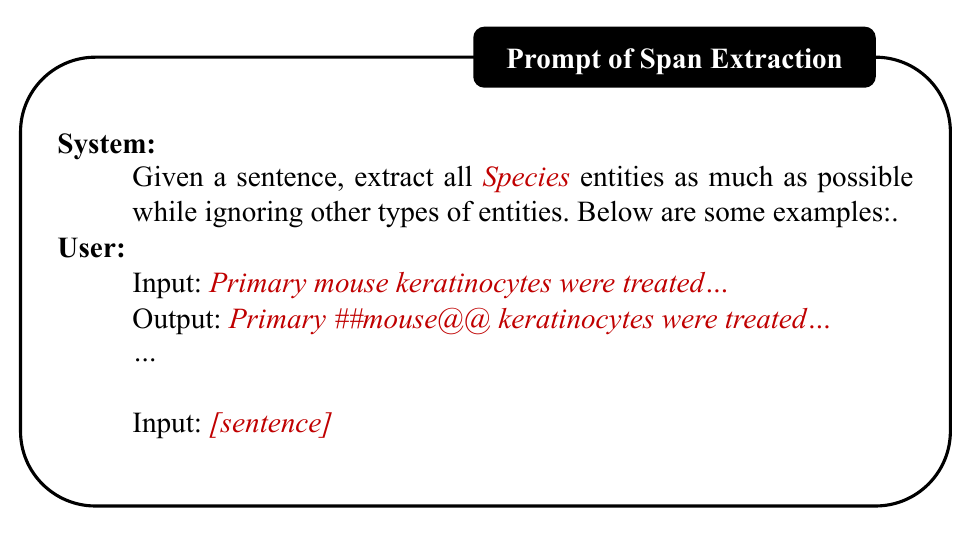}
    \caption{Templete of entity span extraction prompt. The highlighted portions are to be replaced based on the input.}
    \label{fig:prompt1}
\end{figure}
% 加红的部分是要根据输入替换的。
% 我们的方法和
Following GPT-NER\cite{2023_gptner}, we use special markers such as '@@\#\#' to indicate entity positions. For instance, searching for entities of the "\textit{Chemicals}" category in the input text "\textit{A common feature of these proteins is involvement with heterochromatin and/or transcriptional repression}" can be converted into the task of generating the text sequence "\textit{A common feature of these @@proteins\#\# is involvement with heterochromatin and/or transcriptional repression}."

When extracting entities in LLM, the entity type will be specified in the input prompt. For datasets containing multiple entity types, LLM should extract one type of entity at a time and then merge the obtained spans. Figure 1 shows the prompt used for extracting entity spans.
% 图一展示了抽取entity span 用到的 prompt。

\subsection{External knowledge Retriving}

% 在获得了实体的边界后，用 LLM 来判断span的类别，在判断某一个实体span类别的时候，我们会获取和这个span相关的外部知识，启发LLM做出更精准的判断。

% [一旦我们确定了实体的span，下一步要将每个实体span分类到特定的类别中。我们的实体类型预测（ETP）模块采用了大型语言模型（LLM），即GPT-3.5-turbo，用于实体分类。鉴于生物医学文本中领域特定知识的重要性，我们进一步地将一个外部知识库集成到这个模型中。我们将当前的实体跨度以及相关的外部知识一同输入到LLM中。通过这样做，我们旨在激发LLM做出更精确的判断，提高实体分类的准确性。]

Once spans of entities are identified, the subsequent step is to classify each entity span into specific category. Our Entity Type Prediction (ETP) module employs the LLM for category prediction. In view of the importance of domain-specific knowledge in biomedical text, we take a step further by integrating an external knowledge repository into this model. We input both the current entity spans and pertinent external knowledge together to the LLM. By doing so, we aim to inspire the LLM to make more precise judgments and enhancing the accuracy of entity category prediction.

% 在获取数据并构建知识库之后，我们继续进行进一步的指令调优。典型的指令调优示例由三个组成部分组成：指令、输入和输出。指令向模型提供了关于应该执行的任务的信息，输入代表了模型需要处理的数据，输出是基于指令和输入生成的模型结果。这三个组成部分结合在一起，用于训练和微调模型，使其能够执行特定任务。

% 本研究的创新之处在于输入内容包含了外部知识，如图***所示，展示了改进前后的提示内容。

% After obtaining the data and constructing the knowledge base, we proceed with instruction tuning, enabling it to perform specific tasks. A typical instruction-tuning example consists of instructions, input, and output. Instructions provide the model with information about the task it should perform, input represents the data the model needs to process, and output is the result generated by the model based on the instructions and input. 
% The innovation in this study lies in the inclusion of external knowledge within the input content, as depicted in Figure ***, showcasing the prompt content before and after improvement.

% \textbf{External knowledge Retriving}
% 我们采用 UMLS 当作实体类别判断的外部知识库，这个知识库是一个包含了<实体名称，实体类别> 的字典，定义为 KG=「（name, type）」,
% 外部知识的获取可以看作是一个检索问题，query 是实体 span，key 是UMLS字典中实体的名称，返回的是和 query 最相似的 k 个实体以及对应的实体类别：
% 这里的相似性我们用实体name的相似度来衡量，具体做法是对 query name 和 key name 用 sapbert 进行编码，然后计算这两个向量的 cosine 相似度。

% [上述外部知识库是使用统一医学语言系统(UMLS)构建的。该知识库可以描述为一个包含一对<实体名称，实体类别>的字典，形式化表示为KG = "(名称，类型)"。该文将获取外部知识的过程抽象为一个检索任务，以实体跨度作为查询，以词典中实体的名称作为关键字。我们的目标是从字典中检索与查询最相似的前k个实体，以及它们对应的实体类别。使用SapBERT对查询名称和键名进行编码，然后计算它们之间的余弦距离。余弦距离作为实体名称相似度的度量。]

The previously mentioned external knowledge base has been constructed using the Unified Medical Language System (UMLS). This knowledge base can be described as a dictionary containing pairs of $\left< entity\,\,name,\,\,entity\,\,category \right>$, formally denoted as $KG="\left( name,\,\,type \right)"$. We conceptualize the process of acquiring external knowledge as a retrieval task, in which the entity span serves as the query, while the names of entities in dictionary are regarded as the keys. Our objective here is to retrieve the top $k$ most similar entities to the query from the dictionary, along with their corresponding entity categories. We employ SapBERT to encode both the query names and the key names, then compute of the cosine distance between them. This cosine distance then acts as a measure of similarity for the entity names.

% \textbf{Unsupervised dictionary refine}

% 在对实体span进行外部知识检索的过程中，我们发现 topk 个实体中，某些实体出现的频率非常高，即使这个实体和 query 的类别不同
% 我们提出了一个无监督的方法对 UMLS knowledge base 进行升级, 剔除了一些会对结果造成影响的实体

\subsection{Knowledge infused category infer}
\begin{figure}[hbp]
    \centering
    \includegraphics[width=1\columnwidth]{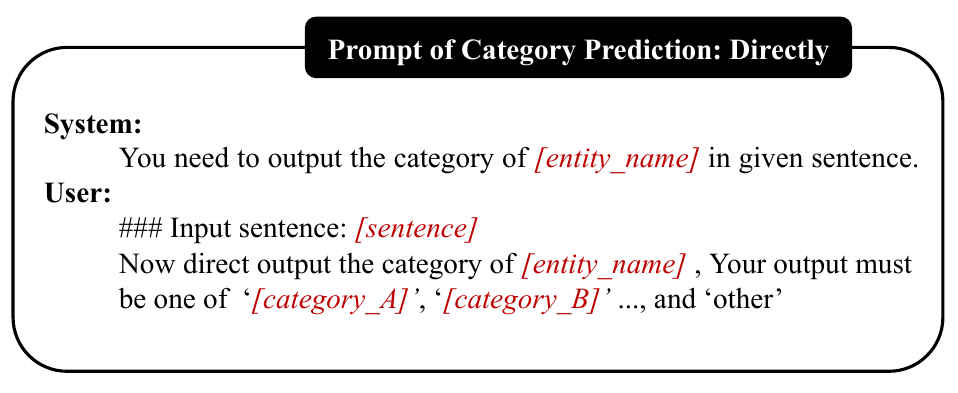}
    \caption{Templete of category prediction prompt. This template allows the LLM to directly determine the entity's category}
    \label{fig:prompt2}
\end{figure}
% 这个模版让 LLM 直接判定实体的类别
% 当实体的span抽取了以后，需要对span的类别进行判断，当然我们可以直接通过 LLM 对 span 进行询问，对应的 prompt 如图2所示。但我们认为，如果有外部知识的介入，模型会做的更好，在 scetion 3.3 中我们已经检索得到了 k 对entity-category pairs，根据这 k 个 knowledge pairs 以及输入的句子和 section3.1 抽取得到的实体，我们可以构建 prompt 模版如图3所示

% 在 section3.1 中，获得实体 span 的时候我们指定了实体的类别，所以理论上可以对实体的类别有一个初步的判断。实际上这样做效果不是很理想。所以我们的方法会对实体 category 有一个重新的判断，在判断的时候会告诉模型实体类别的范围，启发模型做思考。并且设定了一个 'other' 类别，给 LLM 一个纠错的机会。这里的 prompt 如图 2 所示

In section 3.2, when obtaining entity spans, we specify the entity's category, theoretically providing a preliminary judgment of the entity's category. However, in practice, this approach does not yield ideal results. Therefore, our method repredict the entity category and during this judgment process, it informs the model about the range of entity categories. Additionally, we introduce an 'other' category to provide the LLM with a correction opportunity. The prompt for this is illustrated in Figure \ref{fig:prompt2}.

% After extracting the entity span, it's necessary to determine the span's category. While we can directly query the span through the LLM, as illustrated in Figure \ref{fig:prompt2}, we believe that the model can perform better with the introduction of external knowledge. 

% 为了实体类别的判断能更进一步
In order to further improve entity category determination, in section 3.3, we have retrieved k entity-category pairs. Based on these k knowledge pairs, along with the input sentence and the entities extracted in section 3.2, we can construct a prompt template as shown in Figure \ref{fig:prompt2}.

\begin{figure}[hbp]
    \centering
    \includegraphics[width=1\columnwidth]{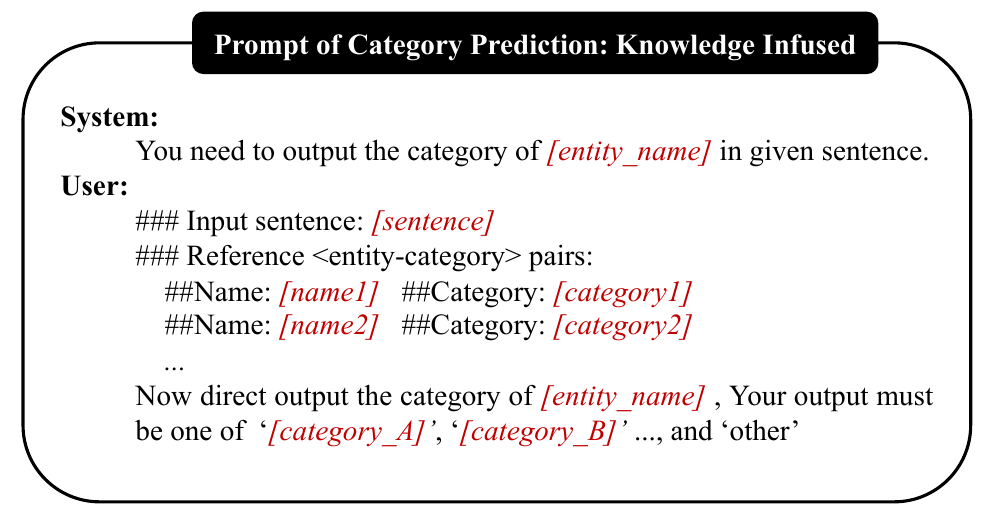}
    \caption{Prompt templete of category prediction with knowledge infused.}
    \label{fig:prompt3}
\end{figure}
% 大纲：
%   首先介绍下数据集，目前只有 craft，别的后面再添加
%   然后是实验细节，比如说用的是 chatgpt 什么版本之类的
%   然后介绍一下 baseline
%   之后是实验结果，在主要的数据集（craft）上分数有多少，比别的方法好多少
%   消融实验，我们的改进对模型有多少提升，例如加入了外部知识提升了多少
%   展示一些例子或者再做一些分析。。。

\section{Experiments}

\begin{table*}[htbp] \small
  \centering
  \begin{tabular}{l|P{9mm}P{9mm}P{9mm}|P{9mm}P{9mm}P{9mm}|P{9mm}P{9mm}P{9mm}}
    \toprule
     \multirow{2}{*}{Model} & \multicolumn{3}{c|}{Species} & \multicolumn{3}{c|}{Gene/Protein} & \multicolumn{3}{c}{Chemical} \\
                & Pre & Rec & F1 & Pre & Rec & F1 & Pre & Rec & F1  \\ \midrule
        \multicolumn{10}{c}{Zero / One - shot} \\ \midrule
        % GPT direct                  & 31.22 & 54.33 & 39.65 & 43.44 & 49.30 & 46.18 & 16.41 & 12.22 & 14.01 \\
        GPTNER-RR                    & 32.60 & 59.05 & 42.01 & 54.54 & 86.91 & 66.90 &  6.71 & 23.33 & 10.42 \\
        \quad + ReType-GPT           & 83.87 & 61.41 & 70.90 & 52.83 & 91.16 & 66.89 & 45.94 & 18.88 & 26.77 \\
        \quad + ReType-KG+VOTE       & 82.97 & 61.41 & 70.59 & 66.16 & 60.93 & 63.43 & 36.00 & 10.00 & 15.65 \\
        \quad + ReType-KG+GPT        & 86.02 & 62.99 & 72.72 & 55.71 & 90.69 & 69.02 & 57.57 & 21.11 & 30.89 \\
          \midrule
        \multicolumn{10}{c}{Cross-Dataset Adaptation} \\ \midrule
        UniversalNER-7B              & 100.0 & 47.24 & 64.17 & 65.50 & 60.93 & 63.13 & 60.00 & 23.33 & 33.60 \\
        \quad + ReType-GPT           & 100.0 & 44.88 & 61.95 & 61.21 & 60.93 & 61.07 & 65.21 & 16.66 & 26.54 \\
        \quad + ReType-KG+VOTE       & 95.23 & 47.24 & 63.15 & 74.62 & 46.51 & 57.30 & 50.00 &  6.66 & 11.76 \\
        \quad + ReType-KG+GPT        & 100.0 & 47.24 & 64.17 & 64.03 & 60.46 & 62.20 & 58.62 & 18.88 & 28.57 \\
        DMNER                       & 91.17 & 48.81 & 63.58 & 63.79 & 34.41 & 44.71 & 57.57 & 10.00 & 17.01 \\
        HUNER                       & 98.51 & 73.83 & 84.40 & 59.67 & 65.98 & 62.66 & 53.56 & 35.85 & 42.95 \\
        % tmChem                      &   -   &   -   &   -   &   -   &   -   &   -   & 49.74 & 31.43 & 38.52 \\
        % GNormPlus                   & 87.03 & 69.51 & 77.29 & 65.03 & 40.55 & 49.95 &   -   &   -   &   -   \\

    \bottomrule
  \end{tabular}
  \caption{Results on Craft Dataset (200).} \label{tab:craft}
\end{table*}

\subsection{BioNER Datasets}
% 我们的实验使用了 CRFFT 数据集，CRAFT 数据集包含了200个样本，有三种类别的实体，分别是 Species，Chemical，Gene/Protein。
% [在我们的实验中，我们使用了CRAFT数据集。本文选用的数据集包含200个样本，涵盖了三个不同的实体类别，分别是物种（Species）、化学物质（Chemical）和基因/蛋白质（Gene/Protein）。]

In our experiments, we utilize the CRAFT\cite{2017_craft} dataset.  The dataset chosen for our study comprises 200 samples, encompassing three distinct entity categories: Species, Chemical, and Gene/Protein.

\subsection{Implementation Details}
In the experiment, we used GPT-3.5 as the LLM to perform entity extraction and category determination. In the step of extracting entity spans, we randomly sampled two samples from the craft dataset as examples for in-context learning. We randomly sampled 500,000 <name, category> pairs from UMLS\cite{2004_umls} as an external knowledge base.
% In the experiment，我们使用 GPT-3.5 作为 LLM 进行实体抽取和类别的判断。在实体span抽取的步骤中，我们随机从craft数据集中采样了两个样本当作 in-context learning 的样例。我们从 UMLS 随机采样 500k 个<name, category> 对当作外部知识库。

\subsection{Baselines}
% 基于大模型 unsupervised 的 baseline 包括:
The unsupervised baseline based on LLM includes:

\begin{itemize}
    \item GPTNER-RR:
    RR is short for Random Retrival. This version of the GPTNER\cite{2023_gptner} model randomly selects a sample as input for one shot inference, while other versions of GPTNER use KNN retrieve similar examples from the training data. Therefore, it cannot be considered one-shot learning.
    % 这个版本的 GPTNER 模型随意选择了一个样本当作做LLM in context-learning 的输入，GPTNER 的其他版本从训练数据中检索了样例，不算是 one-shot learning，所以不参与对比。
    \item ReType-GPT: After extracting the entity span, we then allow GPT to independently determine the category of the span again.
    % 在抽取了实体 span 后，再让 GPT 单独对 span 类别进行判断
    \item ReType-Vote(KG):
    Determine the category of the entity to be predicted by conducting a category voting process based on the categories of similar entities obtained from an external knowledge base.
    % 用外部知识库获取了的类似的实体，参考这些实体的类别投票决定待预测实体的类别。
    \item ReType-GPT3.5(KG):
    Information about similar entities has been obtained from an external knowledge base, and GPT-3.5 is used in conjunction with this information to determine the category of the entity.
    % 从外部知识库获取了类似的实体信息，并且用GPT3.5结合这些来判断实体的类别。
\end{itemize}

% Cross-dataset adaption baseline 有使用其他的生物医学实体数据集训练，然后在 craft 数据集上进行评估。include：
The cross-dataset adaption baseline uses other biomedical entity datasets for training and is then evaluated on the craft dataset. It includes:
\begin{itemize}
    \item UniversalNER-7B:
    A LLAMA based language model, adopt suprvised finetuning on general domain NER datasets, which includes 5 biomedical datasets.
    % 是基于 LLAMA 的语言模型，采用了通用领域的NER数据集进行SFT，其中包含5个biomedical dataset
    \item DMNER\cite{2023_dmner}
    The entity extraction module of DMNER utilizes MRC\cite{2019_unified-MRC} and was trained on five datasets enhanced with ChatGPT. The entity matching module also uses the UMLS dictionary and sapbert.
    % DMNER 的实体抽取模块使用的是 mrc，用chatgpt增强过的5个数据集进行训练，实体匹配模块用到的字典也是 UMLS。
    \item HUNER\cite{2020_huner}:
    The model utilizes the LSTM-CRF framework and was trained on 20 other biomedical entity datasets.
    % 模型采用的是 LSTM-CRF 框架，在20个其他生物医学实体数据集上进行训练
    % \item
\end{itemize}

\subsection{Results and Ablation Study}

\begin{table}[htbp] \small
  \centering
  \begin{tabular}{l|p{10mm}p{10mm}p{10mm}}\toprule
        method           &  Pre  &  Rec  &  F1    \\ \midrule
        % \multicolumn{4}{c}{Overall performance}          \\ \midrule
        VOTE-KG (50k)     & 65.43 & 44.67 & 53.09   \\
        VOTE-KG (100k)    & 70.21 & 45.83 & 55.46   \\
        VOTE-KG (500k)    & 68.76 & 50.46 & 57.21   \\
        GPT-KG (50k)      & 61.80 & 66.66 & 64.14   \\
        GPT-KG (100k)     & 62.01 & 66.89 & 64.36   \\
        GPT-KG (500k)     & 61.74 & 68.05 & 64.76   \\
 \bottomrule
  \end{tabular}
   \caption{On the CRAFT dataset, the impact of knowledge retrieval from UMLS knowledge bases of varying sizes (50k, 100k and 500k) on the results. The GPT method allows the LLM to determine the entity's category, while the VOTE method directly conducts a vote based on the retrieved categories.} \label{tab:aba1}
\end{table}
% 在craft数据集上，引入知识判断实体类别时，不同大小的外部知识库，对于不同模型的影响。
% GPT 方法让LLM决定实体的类别，VOTE 方法直接对检索得到的类别进行投票。
% On the CRAFT dataset, the impact of introducing knowledge for entity category determination using external knowledge bases of varying sizes on different models.
% 从表一可知，直接使用 GPTNER 进行实体的判断效果很差，CRAFT 数据集有三种类型的实体类别，所以实体有很大的概率被误判
From Table \ref{tab:craft}, it is evident that using GPTNER directly for entity classification yields poor results. The CRAFT dataset consists of three different types of entity categories, which significantly increases the likelihood of misclassification.
% 当使用 GPT3.5 重新对实体类别进行评估后，ReType-GPT 的效果提升明显，说明模型在第一步对于实体类别的判断非常模糊
When reevaluating entity categories using GPT-3.5, ReType-GPT shows a significant improvement in performance, indicating that the model's initial judgment of entity categories in the first step was quite ambiguous.
% 除了用 LLM 判断类别，通过外部知识检索的方法 ReType-KG+VOTE 来确定实体的类别效果也不错
In addition to using LLM for category determination, the method of retrieving external knowledge, ReType-KG+VOTE, also performs well in determining the entity's category.
% 效果最好的是 ReType-KG+GPT，在 CRAFT 数据集 gene 类别上，相比于 ReType-GPT F1 提升了 2.27，相比于 ReType-KG+VOTE 效果提升了 5.59，分别说明了引入外部知识，以及用LLM决定实体类别的有效性。
The best-performing approach is ReType-KG+GPT. Specifically, on the CRAFT dataset, in the gene category, it exhibits a 2.27 F1 score improvement compared to ReType-GPT and a 5.59 F1 score improvement compared to ReType-KG+VOTE. These results underscore the effectiveness of incorporating external knowledge and utilizing LLM for entity category determination.

% 在表1上面，我们研究了不同的外部知识库知识量对于 knowlwdge enhanced 方法的影响。直接 vote 的方法，对于外部知识库的数量比较敏感。而采用 GPT 的方法，即使知识库的数量下降了10倍，整体性能也只下降了0.62，这说明了用 LLM 来结合外部知识有更好的鲁棒性。
In Table \ref{tab:aba1}, we examined the impact of the knowledge quantity from different external knowledge bases on knowledge-enhanced methods. The direct vote method is sensitive to the quantity of external knowledge bases. In contrast, the GPT-based method only experiences a 0.62 drop in overall performance even when the knowledge base quantity is reduced by a factor of 10. This demonstrates that using LLM in conjunction with external knowledge offers better robustness.

\subsection{Case Study}

% 对案例进行分析，我们可以看到，cot2的性能与其他方法相比都有一定提升。下面，本文根据Craft数据集上的结果，对性能提升的原因进行相对直观的分析。

% 对案例进行分析，我们可以看到，cot2的性能与其他方法相比都有一定提升。下面，本文根据Craft数据集上的结果，对性能提升的原因进行相对直观的分析。
In this case study, we conducted a comparative analysis of different methods on the Craft dataset, revealing that the cot2 method demonstrates enhanced performance compared to alternative approaches. Below, we provide a nuanced analysis of the underlying factors contributing to this performance improvement:

% 首先，使用GETNER模型可能会导致对并非真正生物医学实体的范围进行错误识别。ETP模块通过将NER任务分为两个步骤，有助于解决BED模块在识别实体范围时引入的错误。例如，在附录的示例1中，GPTNER错误地将“in”识别为生物医学实体。相比之下，其他三种方法将其分类为“其他”，从而在预测该词的结构时进行了纠正。
% 首先，使用GETNER模型可能导致对span的错误识别，这些span不是真正的生物医学实体。ETP模块通过将NER任务拆分为两个步骤，帮助解决BED模块在识别实体跨度时引入的错误。例如，在附录中的示例1中，GPTNER错误地将“in”标识为生物医学实体。相比之下，其他三种方法将其归类为“其他”，从而为该词的预测结构提供了一种纠正机制。

Firstly, using the GETNER model may lead to incorrect recognition of spans that are not truly biomedical entities. The ETP module, by splitting the NER task into two steps, helps address errors introduced by the BED module in identifying entity spans. For instance, in Example 1 in the appendix, GPTNER incorrectly identifies "in" as a biomedical entity. In contrast, the other three methods classify it as "other," thereby offering a corrective mechanism for the prediction.

% 其次，随机检索方法缺乏外部领域知识的支持，对专业医学文本的分类标准不够明确。因此，与包含外部知识的方法相比，它更容易被错误分类。如附录中的示例2所示，最初识别的实体“t抗原”属于与病毒基因组相关的蛋白质类别\cite{1979_T-antigen}，证明使用直接LLM预测精确捕获具有挑战性。这种专业领域知识的复杂性很难通过直接的LLM预测来捕捉。同样的问题也出现在对 Example 3中的实体"thymine–adenine"和Example 4中的实体”Mcoln 1“ 的预测过程中。
Secondly, the Random Retrieval method lacks the support of external domain knowledge and lacks clarity regarding the criteria for classifying professional medical texts. Consequently, compared to approaches that incorporate external knowledge, it is more susceptible to misclassification. As demonstrated in Example 2 in the appendix, the initially identified entity 'T-antigen,' which belongs to a category of proteins associated with viral genomes\cite{1979_T-antigen}, proves challenging to precisely capture using direct LLM prediction. The intricacies of such specialized domain knowledge are difficult to capture through direct LLM predictions. The same situation occurs during the prediction of the entity "thymine-Adenine" in Example 3 and "Mcoln 1" in Example 4.

% 与 '投票' 方法不同，它完全依赖实体相似性进行分类，而数据库通常在数据类别、分布和粒度上存在差异。在处理多样性的测试数据集时，单一知识库很难保持一致的高性能。例如，在附录的示例3中，像 'Mcm4/6/7' 这样的实体在 UMLS 中有细致的分类，包括 'Gene'、'Gene or Genome' 和 'Genetic Function'，这可能导致最初属于 'Gene' 类别的实体被错误分类为 'other'。而LLM 强大的泛化能力使其能够有效克服这一挑战。
Unlike the 'vote' method, which relies solely on entity similarity for classification, databases often have variations in data categories, distribution, and granularity. When dealing with diverse test datasets, maintaining consistent high performance with a single knowledge base becomes challenging. For instance, in Example 3 in the appendix, entities like 'Mcm4/6/7' have fine-grained categorizations in UMLS, including 'Gene,' 'Gene or Genome,' and 'Genetic Function,' leading to potential misclassification of entities originally in the 'Gene' category as 'other.' While LLM's strong generalization capabilities allow it to effectively overcome this challenge.

% 使用gpt的方法在分类时可以同时考虑上下文内容，而不是进行简单的文本相似度匹配。例如在Example4中，huntingtin既是一种蛋白质，也是一种疾病的名称。LLM对于这样存在歧义的单词，能够更好地完成分类。
Using the GPT method in classification also allows for the simultaneous consideration of contextual information, rather than relying solely on simple text similarity matching. For instance, in Example 4, 'huntingtin' is both a protein and a disease name. LLM performs better in classifying such ambiguous words.

% 这个先不急，后面再写

\section{Conclusion}
In this paper, for BioMedical Named Entity Recognition (NER), we approach the task by first extracting entity spans and then determining entity categories. This method breaks down the NER task, and for entity category determination, we introduce a framework for injecting entity knowledge to address the LLM's lack of domain knowledge. We inject relevant entities obtained from an external knowledge base (UMLS) based on entity representation similarity. Experimental results demonstrate a significant improvement in our approach compared to previous LLM few-shot methods, and our method performs comparably to some supervised methods.
% 在这篇论文中，针对 BioMedical NER ，我们用先抽取实体span，再判断实体类别的方法来，拆解NER任务，并在实体类别的判断时，提出了一个实体知识注入的框架来解决LLM缺乏领域知识的问题，通过实体表示的相似度，我们把从外部知识库（UMLS）检索得到的相关实体注入。实验结果表明我们的方法相比于之前的 LLM few-shot 方法提升明显，并且和一些supervied 的方法比较接近。

% Entries for the entire Anthology, followed by custom entries
\bibliography{main}
\bibliographystyle{acl_natbib}
\newpage

\appendix

\section{Appendix}

% \subsection{Dataset Details}

% % 统计 BEL dictionary entity 数量的 table
% \subsection{BEL dictionary}
\subsection{Case study}
The following shows a set of cases, with each case comprising four sections: the \textbf{\textit{Input sentence}}, the \textbf{\textit{gold entity category table}}, the \textbf{\textit{predicted entity category table}}, and the \textbf{\textit{External knowledge table}}. In the Input sentence, words in red represent the gold span, and highlighted words represent the entity boundaries extracted by GETNER. In the External knowledge table, Reference1 to Reference5 correspond to the top 5 entities closest to the entity to be predicted.

\begin{figure*}[hbp]
    \centering
    \includegraphics[width=2\columnwidth]{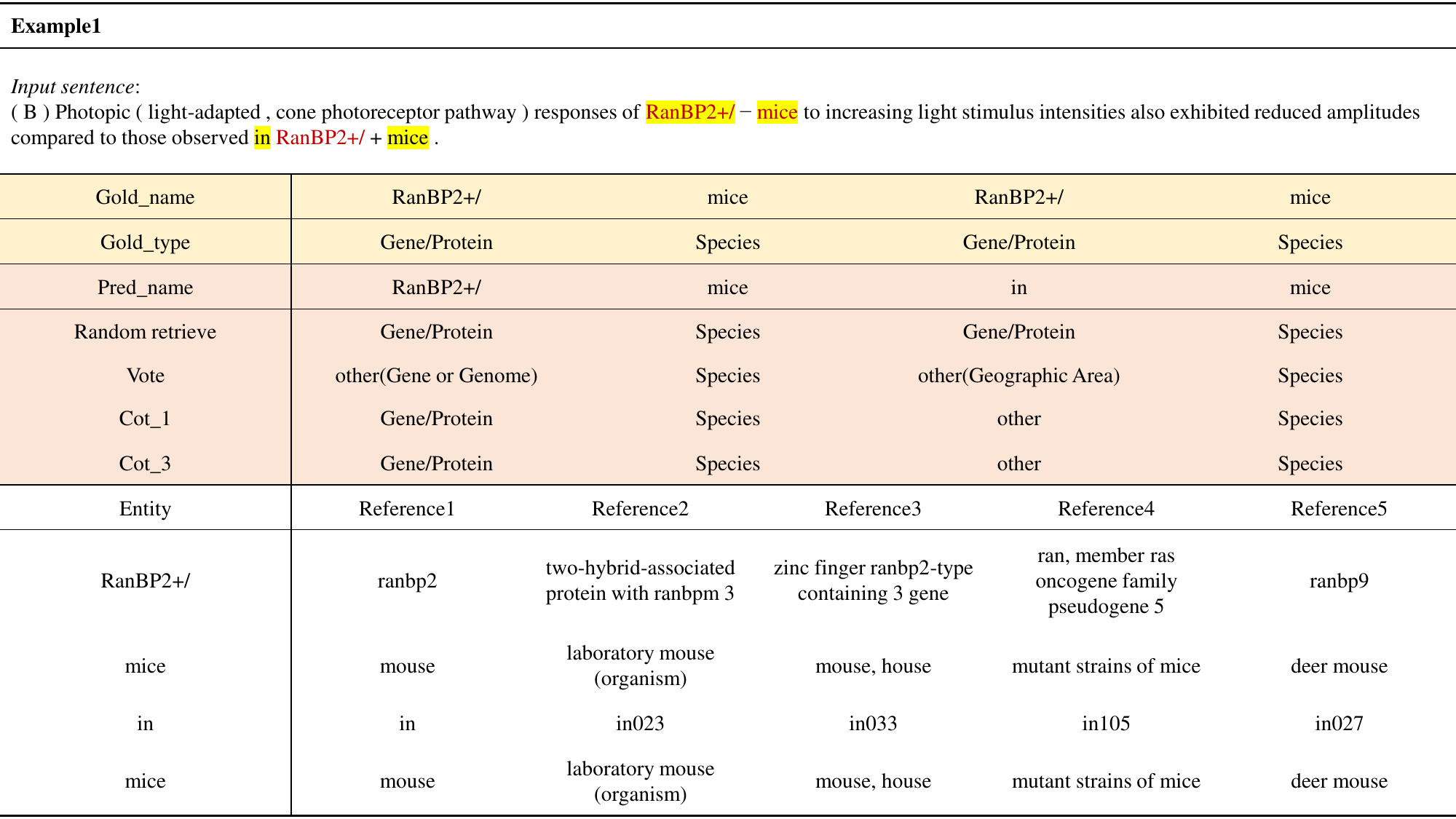}
    \caption{Example 1}
\end{figure*}

\begin{figure*}[hbp]
    \centering
    \includegraphics[width=2\columnwidth]{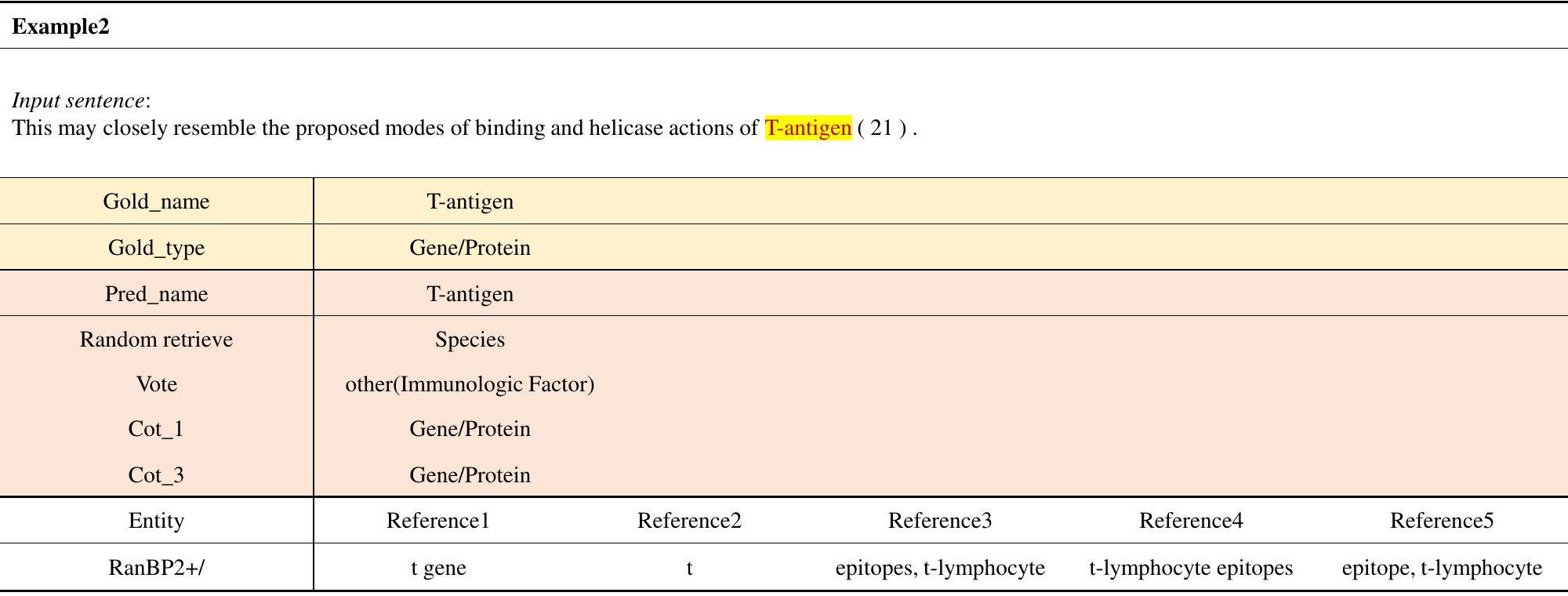}
    \caption{Example 2}
\end{figure*}

\begin{figure*}[hbp]
    \centering
    \includegraphics[width=2\columnwidth]{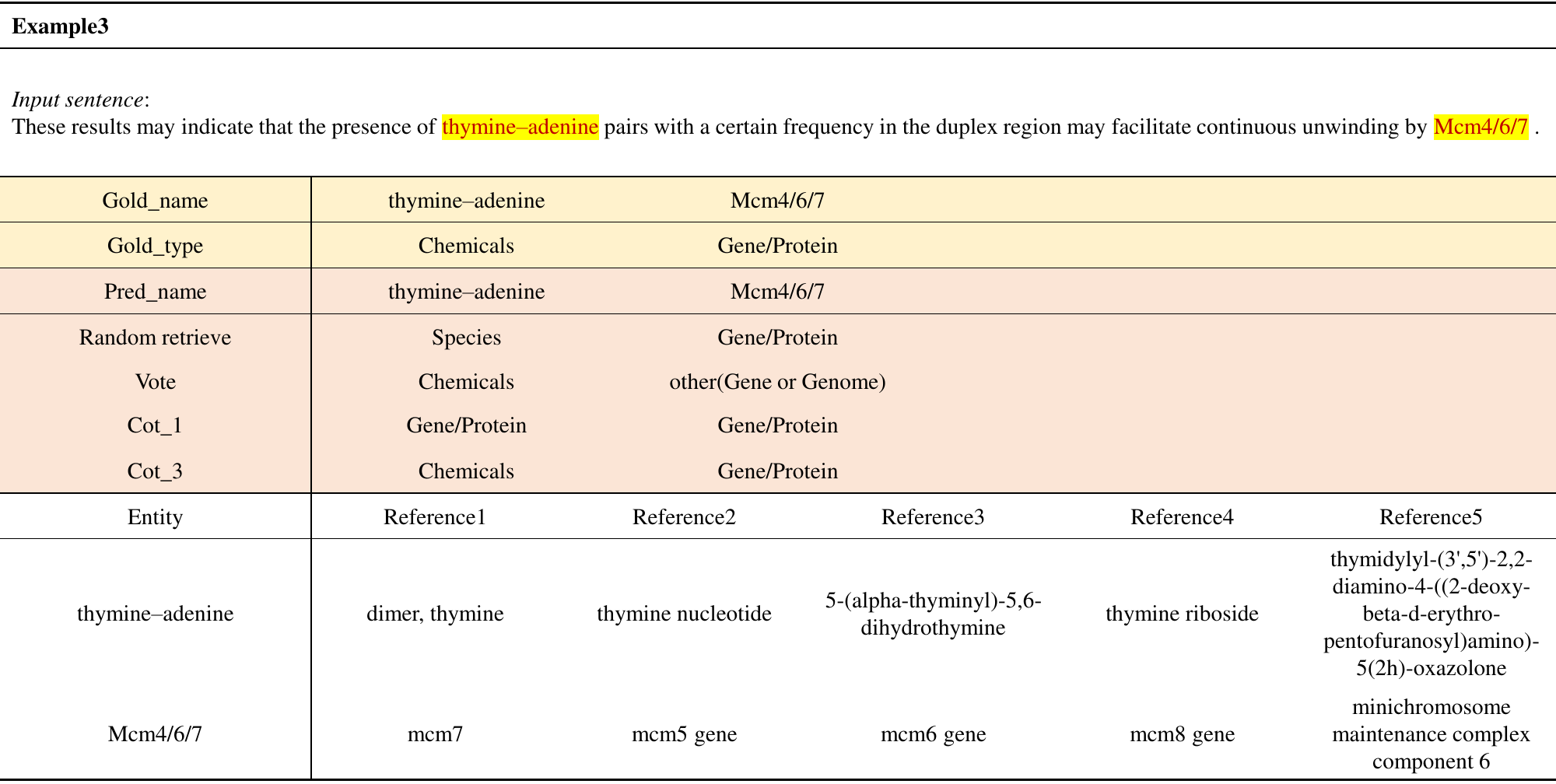}
    \caption{Example 3}
\end{figure*}

\begin{figure*}[hbp]
    \centering
    \includegraphics[width=2\columnwidth]{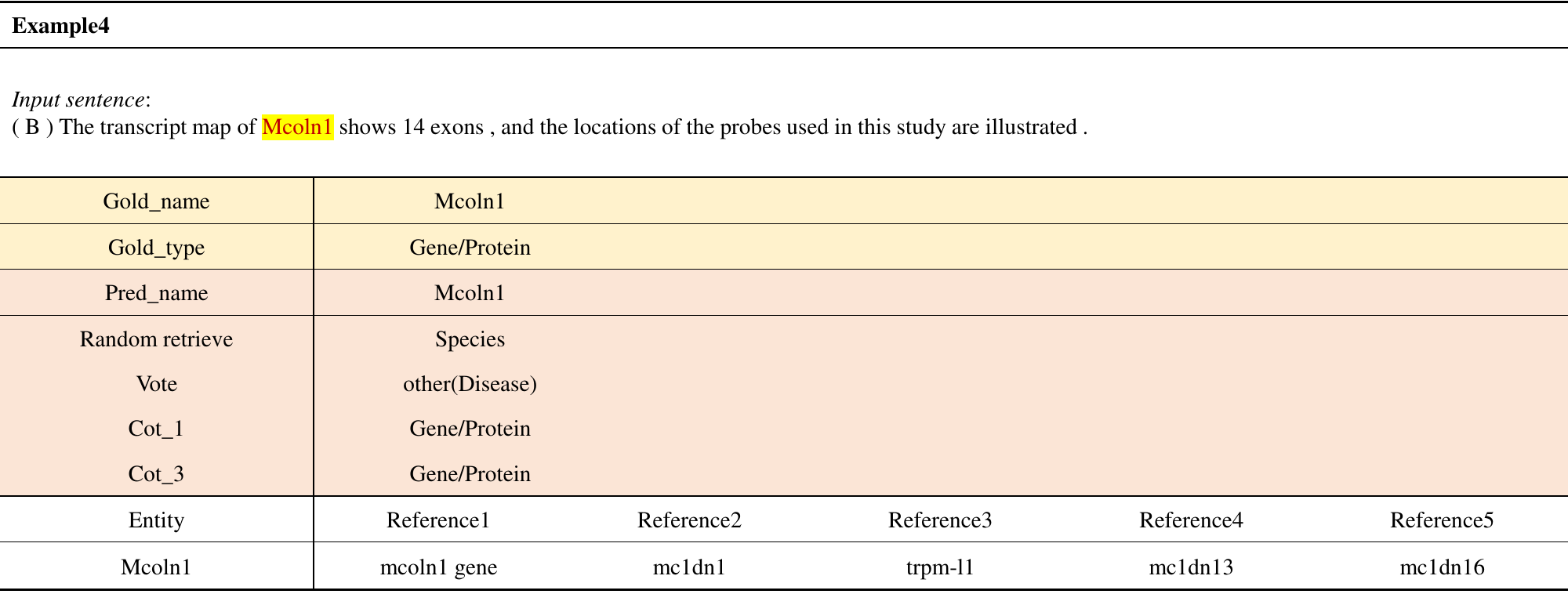}
    \caption{Example 4}
\end{figure*}

\begin{figure*}[hbp]
    \centering
    \includegraphics[width=2\columnwidth]{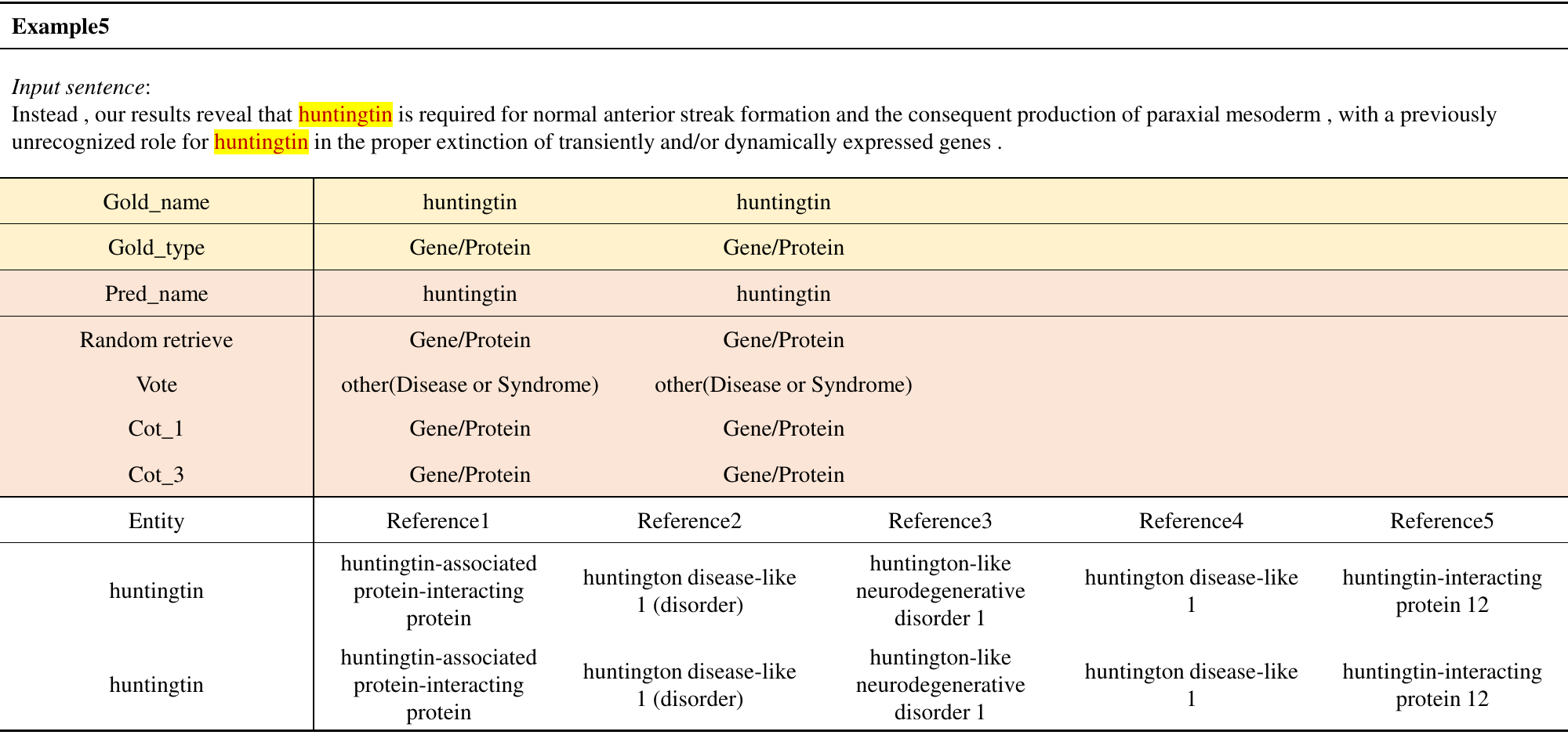}
    \caption{Example 5}
\end{figure*}

% \begin{figure*}
%     \begin{subfigure}{0.8\columnwidth}
%         \centering
%         \includegraphics[width=0.8\linewidth]{./miscs/Example1.pdf}
%     \end{subfigure}
%     % \clearpage % 强制分页
%     \begin{subfigure}{0.8\columnwidth}
%         \centering
%         \includegraphics[width=0.8\linewidth]{./miscs/Example2.pdf}
%     \end{subfigure}
%     % \clearpage % 强制分页
%     \begin{subfigure}{0.8\columnwidth}
%         \centering
%         \includegraphics[width=0.8\linewidth]{./miscs/Example3.pdf}
%     \end{subfigure}
%     \caption{aaa}
%     \label{fig:prompt4}
% \end{figure*}

\label{sec:appendix}

\end{document}